\documentclass[pdflatex,sn-vancouver-num]{sn-jnl}

\usepackage[utf8]{inputenc}
\usepackage[T1]{fontenc}
\usepackage[english]{babel}
\usepackage{graphicx}
\usepackage{multirow}
\usepackage{amsmath,amssymb,amsfonts}
\usepackage{mathrsfs}
\usepackage{xcolor}
\usepackage{textcomp}
\usepackage{manyfoot}
\usepackage{booktabs}
\usepackage{makecell}
\usepackage{array}
\usepackage{float}
\usepackage{placeins}
\usepackage{url}
\usepackage{tikz}
\usetikzlibrary{arrows.meta,positioning}

\hypersetup{hidelinks,hypertexnames=false}

\newenvironment{bmcnolinenumbers}{}{}
\newcommand{\bmctablefont}{\fontsize{7bp}{8bp}\selectfont}
\newcommand{\bmctablenotefont}{\fontsize{6.5bp}{7.5bp}\selectfont}
\begin{document}

\title[Temporal Validation and Public-Health Utility]{Temporal Validation Changes the Apparent Public-Health Utility of Under-Five Mortality Prediction in Bangladesh: A Four-Round DHS Machine-Learning Study}

\author[1]{\fnm{Md Muhtasim Munif} \sur{Fahim}}\email{fahim.stat.ru@gmail.com}
\author[2]{\fnm{M. Monimul} \sur{Huq}}\email{monimul@ru.ac.bd}
\author[2]{\fnm{M.} \sur{Sabiruzzaman}}\email{suza@ru.ac.bd}
\author*[1]{\fnm{Md Rezaul} \sur{Karim}}\email{mrkarim@ru.ac.bd}

\affil[1]{\orgdiv{Data Science Research Lab, Department of Statistics}, \orgname{University of Rajshahi}, \orgaddress{\city{Rajshahi}, \postcode{6205}, \country{Bangladesh}}}
\affil[2]{\orgdiv{Department of Statistics}, \orgname{University of Rajshahi}, \orgaddress{\city{Rajshahi}, \postcode{6205}, \country{Bangladesh}}}

\abstract{\textbf{Background:} Bangladesh has reduced under-five mortality substantially, but preventable deaths remain unevenly distributed across households and divisions. Prediction models based on Demographic and Health Survey data could help planners prioritise follow-up, referral, and resource allocation, but only if their reported performance reflects future public-health use. We quantified how validation design changes apparent under-five mortality prediction performance in Bangladesh.

\textbf{Methods:} We analysed four Bangladesh Demographic and Health Survey rounds (2011, 2014, 2017, and 2022; 33 962 children; 1 290 under-five deaths). The same 26-feature preprocessing pipeline and three model classes were evaluated under four validation regimes: pooled random 80/20, matched-size pooled random 80/20, 2022-only random 80/20, and cross-survey temporal validation trained on 2011+2014, validated and calibrated on 2017, and tested on held-out 2022. The neural model was a 32-unit ELU multilayer perceptron selected by genetic-algorithm neural architecture search. AUROC was estimated with 2 000 bootstrap resamples; screening utility used sensitivity, positive predictive value, and number needed to screen (NNS) at fixed capacity.

\textbf{Results:} Validation regime changed the public-health interpretation of model performance more than model class. For the NAS-derived multilayer perceptron, AUROC ranged from 0.669 under 2022-only random validation to 0.775 under pooled random validation, with a temporal estimate of 0.730. At the top-10 percent temporal screening threshold, NAS predictions identified 152 of 355 observed deaths in 2022 (sensitivity 42.8 percent, positive predictive value 13.2 percent, NNS 7.6). Across validation designs, the same model implied NNS values from 5.6 to 11.0, changing the expected follow-up workload and deaths identified.

\textbf{Conclusions:} In this four-round Bangladesh benchmark, validation-regime choice changed the screening workload and apparent policy value of under-five mortality prediction more than architecture choice. Cross-round temporal validation gives planners a more defensible basis for estimating community-health-worker follow-up, referral demand, and budget scenarios than random-split AUROC alone. DHS-based child-mortality prediction studies should therefore report capacity-based metrics such as sensitivity, positive predictive value, and NNS before such models are used for programme planning or public-policy decisions.}


\maketitle

\section{Background}

Bangladesh has achieved a large reduction in under-five mortality since 1990 \cite{You2015Global,Islam2023,Khan2017}. Earlier analyses of Bangladesh's health transition attribute much of this progress to pluralistic health-service delivery, community-based programmes, female education, and targeted maternal-child health services \cite{Chowdhury2013BangladeshParadox,Arifeen2013Community}. Disparities nevertheless persist across households and regions \cite{Chowdhury2017,Mahumud2021,MNKhan2022}. The broader child-survival literature continues to identify nutrition, maternal education, birth spacing, and access to care as major determinants of survival in low- and middle-income countries \cite{Black2013Maternal,Victora2021Revisiting,Kong2024}.

For health planners, the practical question is not only whether a model can discriminate risk, but whether it gives a realistic estimate of how many children must be contacted to find one child who would otherwise die before age five. A model that appears too efficient may encourage unrealistic staffing, referral, or budget assumptions; a model that appears too weak may discourage targeted follow-up in settings where it could still be useful. Machine learning can model non-linear interactions among child-survival risk factors \cite{Julkunen2025,Rajkomar2019Machine}, but for public-health deployment, discrimination estimated from an internal or random validation set is not enough. Prediction models require external or temporally separated validation before their performance is interpreted as transportable \cite{Moons2012RiskI,Moons2012RiskII,Steyerberg2014Towards}.

Random $k$-fold cross-validation assigns observations to training and test sets without respect to time, allowing future data to inform model training. The result can be overoptimistic test performance when the model is later used in a future population, a form of look-ahead bias \cite{Pohjankukka2017,Roberts2017}. Dataset shift is now a recognised problem in health AI, particularly when models move from development data to deployment settings \cite{Subbaswamy2020Development}. In child-health policy, this validation choice affects screening capacity, expected referral volume, and the number of deaths a programme might expect to identify.

We therefore evaluated the same models under four validation regimes that are common in DHS-machine-learning studies, changing only the validation design while holding the feature set, training pipeline, and model classes constant. The primary analysis used a cross-survey-round temporal split: training on the 2011 and 2014 BDHS rounds, retaining 2017 for architecture selection and calibration, and testing on the held-out 2022 round. This design estimates how a model trained on historical survey data would perform in a later survey population and translates that estimate into screening workload for public-health use.

We used Neural Architecture Search \cite{Zoph2017} via a genetic algorithm \cite{Kefale2025} and domain-informed feature engineering. In the geographic fairness analysis, we observed a division-level predictability gradient: socioeconomic predictors discriminated mortality better in less affluent divisions, while Dhaka and Khulna had the lowest AUROCs (Pearson r = $-$0.63 to $-$0.65, depending on wealth convention). We interpret this cautiously through the epidemiologic transition \cite{Omran2005}: as mortality falls, household socioeconomic indicators may explain a smaller share of residual deaths, and survey-based prediction may become less informative.

\section{Methods}

\subsection{Study Design and Data Sources}

This retrospective cohort analysis used Bangladesh Demographic and Health Survey (BDHS) data from four survey rounds (2011, 2014, 2017, 2022; n = 7 601, 6 779, 8 044, and 11 538 births, respectively). After eligibility restrictions and exclusion of records without survival information, the analytic sample comprised 33 962 births. The primary outcome was under-five mortality, defined as death before age five years and measured from maternal birth-history recall.

\begin{figure}[H]
\centering
\resizebox{0.94\textwidth}{!}{%
\begin{tikzpicture}[
  block/.style={rectangle, draw=black, fill=white, align=left, text width=6.25cm, minimum height=1.25cm, inner sep=5pt, line width=0.35pt, rounded corners=1pt, font=\footnotesize},
  side/.style={rectangle, draw=black, fill=white, align=left, text width=4.55cm, minimum height=1.35cm, inner sep=5pt, line width=0.35pt, rounded corners=1pt, font=\footnotesize},
  link/.style={-{Stealth[length=2.0mm,width=1.5mm]}, draw=black, line width=0.4pt, shorten >=1pt, shorten <=1pt}
]
\node[block] (data) at (0,0) {\textbf{1. BDHS rounds}\\2011, 2014, 2017, 2022; 33\,962 births; 1\,290 deaths};
\node[block] (features) at (0,-1.80) {\textbf{2. Preprocessing and features}\\train-only imputation/scaling; 31 engineered candidates; 26 final inputs};
\node[block] (split) at (0,-3.60) {\textbf{3. Temporal separation}\\train 2011+2014; validate/calibrate 2017; test on held-out 2022};
\node[block] (search) at (0,-5.40) {\textbf{4. GA-NAS search}\\25 candidates over 15 generations; width, activation, dropout, batch normalisation};
\node[block] (fixed) at (0,-7.20) {\textbf{5. Fixed neural architecture}\\26-input, 32-unit ELU MLP; dropout 0.3; no batch normalisation};
\node[block] (models) at (0,-9.00) {\textbf{6. Model and validation benchmark}\\NAS MLP, XGBoost, and logistic regression under standardised preprocessing};
\node[block] (outputs) at (0,-11.20) {\textbf{7. Policy-facing outputs}\\AUROC, calibration, sensitivity, PPV, NNS, DCA; screening workload and referral/budget scenarios};

\node[side] (regimes) at (6.90,-3.60) {\textbf{Validation regimes}\\pooled random; matched-$n$ random; 2022-only random; cross-round temporal};
\node[side] (stress) at (6.90,-8.10) {\textbf{Ablation and sensitivity}\\feature domains; architecture sweeps; censoring; missing data; SHAP/equity};
\node[side] (policy) at (-6.90,-11.20) {\textbf{Public-health use}\\community-health-worker follow-up planning; referral triage; capacity and budget assumptions};

\draw[link] (data) -- (features);
\draw[link] (features) -- (split);
\draw[link] (split) -- (search);
\draw[link] (search) -- (fixed);
\draw[link] (fixed) -- (models);
\draw[link] (models) -- (outputs);
\draw[link] (split) -- (regimes);
\draw[link] (models.east) -- (stress.west);
\draw[link] (stress.south west) -- ++(-0.70,-0.55) |- (outputs.east);
\draw[link] (outputs) -- (policy);
\end{tikzpicture}%
}
\caption{\textbf{Study workflow, temporal validation, and policy-evaluation framework.} The full pipeline starts with four BDHS rounds, applies train-only preprocessing to 26 final features, uses 2011+2014 for model development, reserves 2017 for GA-NAS architecture selection and Platt calibration, and tests all models on held-out 2022 data. Comparator validation regimes, ablation and sensitivity checks, and policy-facing metrics are evaluated under the same preprocessing and model-class framework.}
\label{fig:methodology_framework}
\end{figure}

\begin{figure}[H]
\centering
\includegraphics[width=0.90\textwidth]{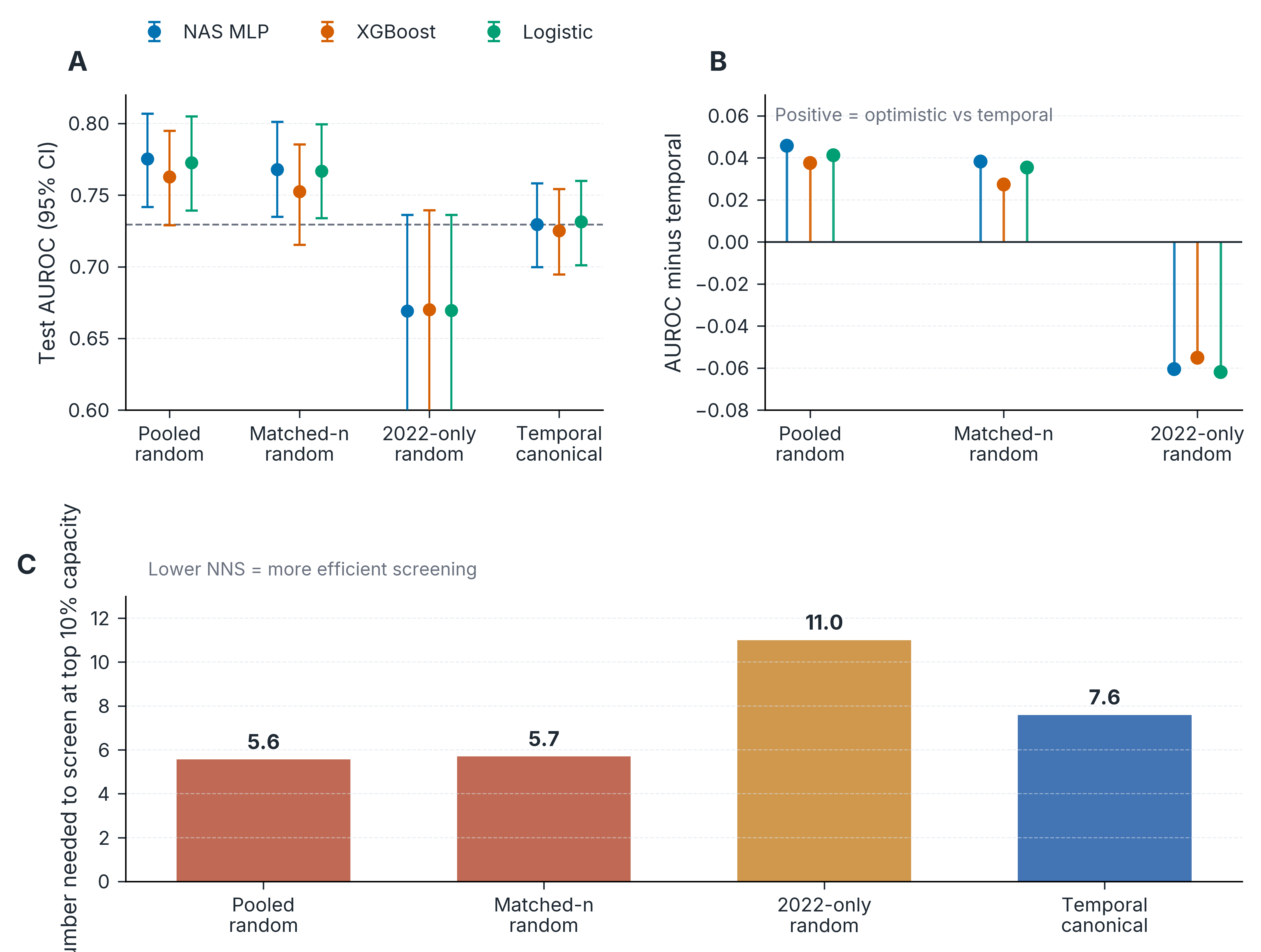}
\caption{\textbf{Validation regime choice changes apparent predictive performance and screening workload.} Panels compare AUROC, displacement from the strict temporal benchmark, and top-10\% NAS screening workload. NNS = number needed to screen. Error bars show 95\% bootstrap confidence intervals from 2 000 resamples.}
\label{fig:validation_regime}
\end{figure}

To quantify the methodological effect of validation design, we evaluated the same feature set and model classes under four regimes. \textbf{(i)} A pooled random 80/20 split across all 2011--2022 records, representing the common practice of mixing survey rounds before splitting. \textbf{(ii)} The same pooled random test set with the training set sub-sampled to match the cross-round temporal training size (n = 14 380), separating validation design from training-sample size. \textbf{(iii)} A 2022-only random 80/20 split, representing the common practice of training and testing on a single DHS round \cite{Rahman2025,Mangold2021}. \textbf{(iv)} Cross-survey-round temporal validation: training on 2011 and 2014, retaining 2017 for architecture selection and calibration, and testing on the held-out 2022 round. All preprocessing was fit inside each training split only: missing values were replaced by training-set medians and continuous features were standardised using training-set means and standard deviations.

The cross-round temporal design is the deployment-relevant reference: it asks how a model developed from earlier survey rounds performs when applied to a future BDHS population. The full study spans eleven years (2011--2022), with a five-year prospective gap between the 2017 model-selection round and the 2022 test round. We intentionally exclude 2017 from the training set of the temporal regime so that 2017 can serve as a clean architecture-selection and calibration round, free of leakage into the deployment test set: had 2017 been included in temporal training, the NAS would have been selected using information correlated with the test distribution, and Platt-scale calibration would no longer be honest out-of-sample. A two-round (2011 + 2014) training set is therefore both the operationally simpler choice and the more conservative one. As a sensitivity, we report in the supplement that a three-round (2011 + 2014 + 2017) temporal training set gives a marginally higher NAS test AUROC on 2022, well within the bootstrap CI of the primary estimate.

Across-round sample independence is supported by the BDHS sampling design \cite{NIPORT2023}: each round draws an independent stratified probability sample of enumeration areas (EAs) from the prevailing Bangladesh Bureau of Statistics census master sample (2011 census frame for the 2011, 2014, and 2017 rounds; 2022 census frame for the 2022 round), with no longitudinal household follow-up. The probability that any individual mother appears in two rounds is well below 1\% and the probability that a given birth record falls within the recall windows of two rounds is smaller still. We treat the cross-round samples as independent at the individual level for temporal-validation purposes, but explicitly acknowledge as a limitation that geographic re-use of EAs across rounds cannot be quantified from the standard BDHS recode files used here (it would require the DHS-Geographic-Encoded files, which are released under restricted access). If residual EA-level non-independence exists, it would inflate apparent temporal-generalisation performance. It would bias the cross-round AUROC \emph{upward} relative to a truly geographically disjoint deployment population. This strengthens our argument that pooled-random and matched-$n$ validation are over-optimistic.

\subsection{Feature Engineering}

Because the BDHS surveys contain high-cardinality variables with non-linear associations with child mortality \cite{Kefale2025}, we did not feed raw survey responses into the models. Instead, predictors were transformed into clinically relevant risk categories identified in previous epidemiologic studies of child survival:

\begin{itemize}
\item \textbf{Maternal Factors:}
  \begin{itemize}
  \item Maternal age in three groups: adolescent ($<$19 years; elevated obstetric risk), optimal (19--35), advanced ($>$35; elevated chromosomal and obstetric risks)
  \item Highest education level attained
  \item Total parity
  \end{itemize}
\item \textbf{Household and Socioeconomic Factors:}
  \begin{itemize}
  \item Wealth quintile derived from DHS-standard PCA asset-index factor score
  \item Urban or rural residence
  \item Administrative division
  \end{itemize}
\item \textbf{Healthcare Utilisation Factors:}
  \begin{itemize}
  \item Antenatal-care adequacy (WHO 4-visit threshold \cite{WHO2016})
  \item Delivery location (home vs.\ facility)
  \item Skilled birth attendance
  \end{itemize}
\item \textbf{Child-Specific Factors:}
  \begin{itemize}
  \item Preceding birth interval categorised as high risk ($<$18 months, maternal depletion), moderate (18--36 months), or low risk ($>$36 months)
  \item Birth order
  \item Perceived birth size as proxy for birth weight
  \end{itemize}
\end{itemize}

This domain-driven transformation reduced more than 50 raw survey variables to 31 engineered candidate features, of which 26 were retained in the final model input. The transformation kept the model input interpretable for clinical and policy review.

\subsection{Neural Architecture Search}

\begin{figure}[H]
\centering
\includegraphics[width=0.90\textwidth]{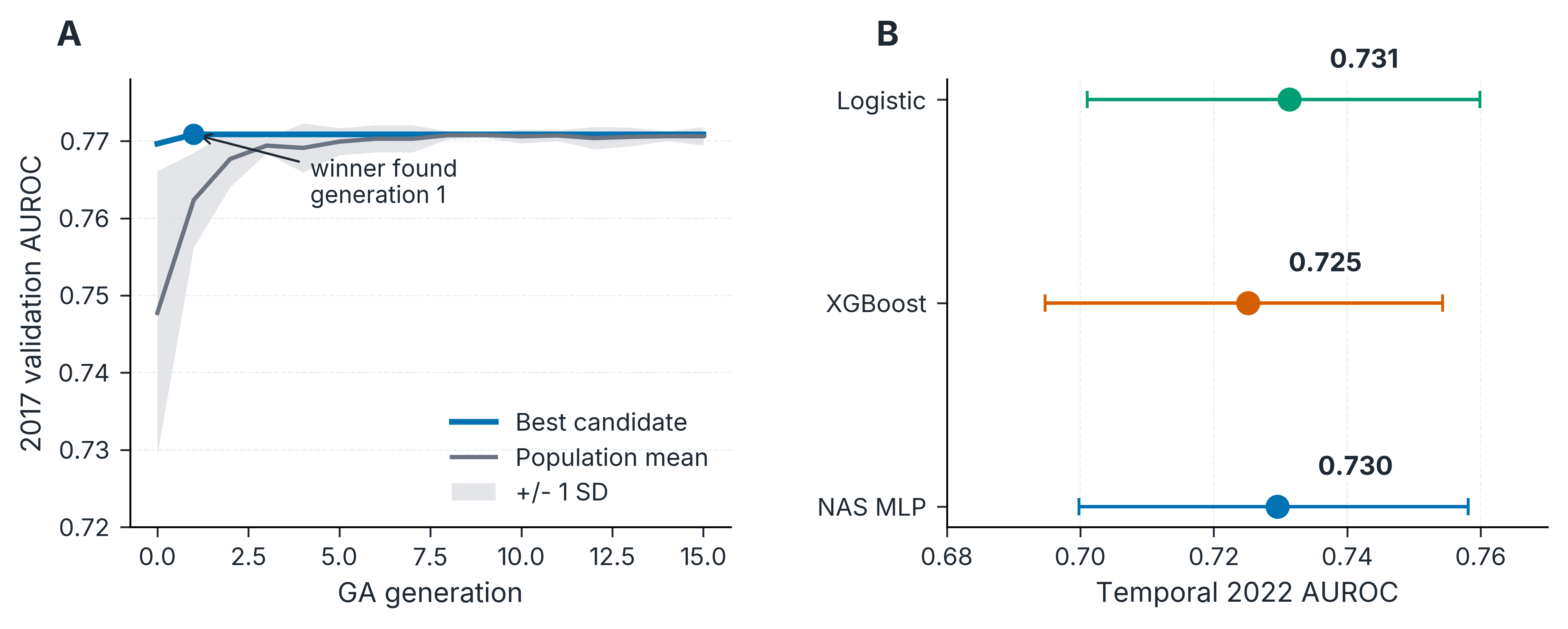}
\caption{\textbf{Genetic-algorithm-driven Neural Architecture Search (GA-NAS).} Panel A shows convergence of the genetic algorithm on the 2017 validation round, and Panel B compares the fixed NAS-derived MLP with XGBoost and logistic regression on the held-out 2022 temporal test set. The search-space and downstream validation workflow are summarised in Figure~\ref{fig:methodology_framework}.}
\label{fig:nas_architecture}
\end{figure}

The neural model class was identified once, by a 15-generation genetic algorithm (population 25; tournament selection $k = 3$; single-point crossover; per-layer mutation on units, activation, dropout, and batch normalisation; elitism of the top 2) over a search space spanning hidden-layer width 32--512, activations \{ReLU, leaky ReLU, ELU\}, dropout 0.0--0.5, batch-normalisation on/off, and 1--4 hidden layers, with candidate fitness scored as validation AUROC on the 2017 BDHS round (train 2011+2014). The search converged within four generations to a 26-input, 32-unit hidden layer with ELU activation, dropout 0.3, and no batch normalisation (Figure~\ref{fig:nas_architecture}); this fixed architecture was then used as the ``neural model'' class in every validation-regime comparison. Figure~\ref{fig:methodology_framework} summarises the GA-NAS search space and the downstream validation, ablation, and policy-metric workflow. We do not claim that this architecture outperforms established tabular baselines. Deep models often do not dominate tree-based or linear methods on structured tabular data \cite{Shwartz-Ziv2022,Grinsztajn2022}; in our temporal benchmark, the GA-v2 MLP, XGBoost, and $\ell_2$-penalised logistic regression had statistically indistinguishable test-set AUROCs. The search was used to define the neural architecture before comparing validation regimes.

\subsection{Comparison Models and Evaluation}

The standardised validation-regime benchmark compared three model classes: the GA-v2-derived 32-unit ELU MLP described above, XGBoost, and $\ell_2$-penalised logistic regression. We restricted the benchmark to these three classes because the substantive question was how much apparent performance changed when the validation regime changed, not whether an exhaustive model zoo could improve AUROC. XGBoost and logistic-regression hyperparameters were fixed at the values previously tuned in the temporal analysis (XGBoost: \texttt{n\_estimators} = 100, \texttt{max\_depth} = 4, \texttt{learning\_rate} = 0.0117, \texttt{subsample} = 0.730, \texttt{colsample\_bytree} = 0.755, \texttt{gamma} = 0.829; logistic regression: $C$ = 0.003, $\ell_2$ penalty). Class weighting was refit within each training split. All AUROCs were reported with 2 000-resample bootstrap 95\% confidence intervals. Reporting was aligned with TRIPOD and TRIPOD+AI principles, and interpretation of prediction-model validity followed PROBAST risk-of-bias concepts \cite{Collins2015TRIPOD,Collins2024,Wolff2019PROBAST}.

\subsection{Additional Analyses}

\begin{itemize}
\item \textbf{Class imbalance.} The positive class is a minority ($\sim$3.8\%). We addressed this through class-weighted loss functions (cost-sensitive learning), which optimised discrimination without synthesising artificial data.

\item \textbf{Calibration.} Calibration was assessed using Brier score and reliability diagrams on the 2022 test set \cite{VanCalster2016Hierarchy,VanCalster2019Calibration}. All three model classes (NAS-MLP, XGBoost, $\ell_2$-logistic) were trained with class-weighted loss to address the 3.8\% positive-class prevalence, which inflates their raw probabilities and yields high pre-calibration Brier scores (NAS 0.151, XGBoost 0.181, logistic 0.158 on the temporal test set). We Platt-scaled all three using their 2017-round predictions as the calibration sample \cite{Platt1999}, restoring all three Brier scores to a tight range (NAS 0.029, XGBoost 0.029, logistic 0.029). Calibrated probabilities are used for Decision Curve Analysis (Additional file 3: Figure S4) and any policy-facing summary that requires probability semantics \cite{Vickers2006Decision}; rank-based screening metrics (top-10\% sensitivity, PPV, NNS) are invariant to monotone recalibration and are reported using the raw scores.

\item \textbf{Geographic equity.} We computed division-level AUROC across the 8 administrative divisions of Bangladesh and Pearson correlation between mean division wealth (2022 test-period convention) and division AUROC.

\item \textbf{Model interpretability.} We computed SHAP values \cite{Lundberg2017} using \texttt{KernelExplainer} so that feature attributions were produced by the same model-agnostic explainer across the NAS-MLP and tree-based baselines. Model-specific explainers (\texttt{DeepExplainer} for neural networks, \texttt{TreeExplainer} for tree ensembles) would not have been directly comparable. The beeswarm visualisation (Figure~\ref{fig:shap_explainability}) and full feature ranking (Additional file 2: Table S7) use a random 3 000-row sample of the 2022 test set (seed = 42; KernelExplainer nsamples = 1 024) with the same 100-row randomly sampled background set for the conditional-expectation reference.

\item \textbf{Validation-regime sensitivity.} The primary sensitivity analysis was the standardised comparison of pooled random, matched-$n$ random, 2022-only random, and cross-survey temporal validation. All AUROCs were reported with 2 000-resample bootstrap confidence intervals. Deployment translation used a fixed top-10\% predicted-risk threshold to reflect constrained screening capacity rather than an unconstrained classification rule.

\item \textbf{Censoring sensitivity.} Because all 2022 test children were under 60 months at interview, we performed a sensitivity analysis restricting alive children to those observed for at least $X$ months at interview ($X \in \{6, 12, 18, 24, 36, 48\}$), retaining all observed deaths. The age-at-death distribution and a stratified-AUROC table are reported in the Supplement.

\item \textbf{Missing-data sensitivity.} Across the tested imputation strategies, median, mode, and zero imputation produced identical XGBoost AUROC (0.686) on the temporal test set; adding binary missing-indicator features did not change discrimination. This sensitivity analysis used a 20-feature XGBoost specification and is reported separately from the primary standardised 26-feature benchmark (temporal XGBoost AUROC 0.725). Downstream models use train-only median imputation as a conservative default.
\end{itemize}

\section{Results}

\subsection{Validation-Regime Comparison}

The main result was that validation regime changed apparent discrimination substantially more than model class. Table~\ref{tab:validation_regime} documents the data composition behind each validation design, while the full numeric benchmark is retained in Additional file 2: Table S1. In every regime the same 26-feature input, train-only preprocessing, model classes, and threshold rule were used; only the validation design changed.

\begin{bmcnolinenumbers}
\begin{table}[!htbp]
\centering
\bmctablefont
\setlength{\tabcolsep}{3pt}
\renewcommand{\arraystretch}{1.12}
\caption{\textbf{Validation-design composition across BDHS rounds.} This table describes the data structure behind each validation design; performance estimates are reported in the Results text, Figure~\ref{fig:validation_regime}, and Additional file 2: Table S1. Pooled random designs mix survey years before splitting, the matched-$n$ design uses the same pooled test set with temporal-size training data, the 2022-only design is a single-round split, and the temporal design is the deployment-style reference.}
\label{tab:validation_regime}
\begin{tabular}{@{}>{\raggedright\arraybackslash}p{2.35cm}
                >{\raggedright\arraybackslash}p{2.05cm}
                >{\raggedright\arraybackslash}p{1.75cm}
                >{\raggedright\arraybackslash}p{2.85cm}
                >{\raggedright\arraybackslash}p{2.75cm}@{}}
\toprule
\textbf{Design} & \textbf{Training rounds} & \textbf{Test round} & \textbf{Training set} & \textbf{Test set} \\
\midrule
Pooled random 80/20 & (2011--2022) & Mixed (2011--2022) & 27\,169 births; 1\,032 deaths (3.8\%) & 6\,793 births; 258 deaths (3.8\%) \\
\addlinespace[0.25em]
Pooled random, matched-$n$ & (2011--2022) & Same pooled test set & 14\,380 births; 546 deaths (3.8\%) & 6\,793 births; 258 deaths (3.8\%) \\
\addlinespace[0.25em]
2022-only random 80/20 & 2022 & 2022 & 9\,230 births; 284 deaths (3.1\%) & 2\,308 births; 71 deaths (3.1\%) \\
\addlinespace[0.25em]
Temporal cross-round & 2011, 2014 & Held-out 2022 & 14\,380 births; 614 deaths (4.3\%) & 11\,538 births; 355 deaths (3.1\%) \\
\bottomrule
\end{tabular}
\begin{tablenotes}
\bmctablenotefont
\item The analytic sample by survey round is 7 601 births/348 deaths in 2011, 6 779/266 in 2014, 8 044/321 in 2017, and 11 538/355 in 2022.
\item This table intentionally does not repeat the AUROC and NNS values plotted in Figure~\ref{fig:validation_regime}; it documents the multivariate design context behind those visual results.
\end{tablenotes}
\end{table}
\end{bmcnolinenumbers}

For the GA-v2 MLP, AUROC ranged from 0.669 (2022-only random) to 0.775 (pooled random), a between-regime spread of 0.106. Relative to the temporal benchmark (0.730), pooled random over-estimated AUROC by 0.046; the matched-$n$ random analysis reduced this to 0.038, showing that most of the gap remained after controlling training-sample size. In contrast, the 2022-only random split under-estimated temporal AUROC by 0.060.

The same direction of bias appeared for XGBoost and logistic regression. Across models within a fixed regime, the maximum AUROC spread was 0.015, whereas the same model varied by approximately 0.09--0.11 across validation regimes (Figure~\ref{fig:validation_regime}A).

The 2022-only under-estimation is not driven by small-sample variance alone. As an auxiliary check we repeated the single-round regime on the 2011 and 2014 BDHS rounds individually, which have \emph{smaller} test samples (n = 1 521 and 1 356, respectively) than 2022-only (n = 2 308). The 2011-only and 2014-only NAS AUROCs were 0.768 [0.704, 0.826] and 0.753 [0.673, 0.831], both \emph{higher} than the cross-round temporal benchmark of 0.730 and lying close to the pooled-random estimate. The single-round bias direction therefore depends on the round chosen: in the higher-mortality earlier rounds, within-round structural signal is strong and single-round random validation \emph{over-states} prospective performance; in the lower-mortality 2022 round, structural signal has weakened and single-round random validation \emph{under-states} prospective performance. The cross-round temporal estimate sits between the earlier-round and 2022-only single-round estimates and is the design-relevant reference for deployment.

To partition the cross-regime AUROC spread into the share due to training-set differences versus the share due to test-set differences, we paired the saved predictions on the intersection of test indices and ran a paired bootstrap of $\Delta$AUROC for every regime pair (Additional file 2: Table S2). The pooled-random and matched-$n$ regimes share the same 6 793-row test set, so their paired comparison isolates the training-set-size effect: for the NAS MLP, $\Delta\text{AUROC} = -0.0075$ [$-0.015$, $0.000$] (p = 0.05), and the equivalent estimates for XGBoost and logistic regression are smaller and non-significant. The temporal and 2022-only regimes share 2 308 test rows; restricted to those shared rows the NAS $\Delta\text{AUROC}$ is $-0.011$ [$-0.065$, $+0.041$] (p = 0.67). The temporal and pooled-random regimes share 2 387 rows; restricted to those rows the NAS $\Delta\text{AUROC}$ is $+0.006$ [$-0.012$, $+0.022$] (p = 0.54). Two implications follow. First, the marginal 0.04--0.10 AUROC differences between regimes in Additional file 2: Table S1 are largely a consequence of which test rows enter each regime---i.e.\ of test-set composition---and therefore should not be interpreted as a property of the trained models. Second, the policy-facing interpretation (NNS, sensitivity at fixed capacity) still differs across regimes because the test-set composition itself differs, which is exactly the methodological point of this paper.

The top-10\% screening results translate these differences into programmatic terms, and Table~\ref{tab:screening_yield} compares the same policy threshold across validation regimes and model classes. Under the temporal reference, the GA-v2 MLP screened 1 154 children and identified 152 of 355 observed deaths at the 10\% capacity point (sensitivity 42.8\%, PPV 13.2\%, NNS 7.6). Pooled random validation made screening look more efficient (NNS 5.6, Figure~\ref{fig:validation_regime}C), while 2022-only random validation made the same model look less efficient (NNS 11.0). At the temporal threshold, XGBoost identified 149/355 deaths (sensitivity 42.0\%, NNS 7.7; exact McNemar p = 0.55 vs.\ NAS), and logistic regression identified 152/355 (sensitivity 42.8\%, NNS 7.6; exact McNemar p = 1.00). These between-model differences are not statistically significant. The prospective ROC curves for these three models are shown in Additional file 3: Figure S3.

\begin{bmcnolinenumbers}
\begin{table}[!htbp]
\centering
\bmctablefont
\setlength{\tabcolsep}{3pt}
\caption{\textbf{Screening utility at 10\% program capacity.} The same top-10\% predicted-risk threshold is applied within each validation regime and model class to show how validation design changes policy-facing workload and yield.}
\label{tab:screening_yield}
\begin{tabular*}{\textwidth}{@{}>{\raggedright\arraybackslash}p{2.15cm}>{\raggedright\arraybackslash}p{1.2cm}@{\extracolsep{\fill}}rrrrrrr@{}}
\toprule
\textbf{Design} & \textbf{Model} & \textbf{$n_{\mathrm{test}}$} & \textbf{Deaths} & \textbf{Screened} & \textbf{TP} & \textbf{Sens.} & \textbf{PPV} & \textbf{NNS} \\
\midrule
Pooled random & NAS MLP & 6793 & 258 & 679 & 122 & 47.3\% & 18.0\% & 5.6 \\
Pooled random & XGBoost & 6793 & 258 & 679 & 120 & 46.5\% & 17.7\% & 5.7 \\
Pooled random & Logistic & 6793 & 258 & 679 & 119 & 46.1\% & 17.5\% & 5.7 \\
Matched-$n$ pooled & NAS MLP & 6793 & 258 & 679 & 119 & 46.1\% & 17.5\% & 5.7 \\
Matched-$n$ pooled & XGBoost & 6793 & 258 & 679 & 117 & 45.3\% & 17.2\% & 5.8 \\
Matched-$n$ pooled & Logistic & 6793 & 258 & 679 & 120 & 46.5\% & 17.7\% & 5.7 \\
2022-only & NAS MLP & 2308 & 71 & 231 & 21 & 29.6\% & 9.1\% & 11.0 \\
2022-only & XGBoost & 2308 & 71 & 231 & 24 & 33.8\% & 10.4\% & 9.6 \\
2022-only & Logistic & 2308 & 71 & 231 & 21 & 29.6\% & 9.1\% & 11.0 \\
Temporal & NAS MLP & 11538 & 355 & 1154 & 152 & 42.8\% & 13.2\% & 7.6 \\
Temporal & XGBoost & 11538 & 355 & 1154 & 149 & 42.0\% & 12.9\% & 7.7 \\
Temporal & Logistic & 11538 & 355 & 1154 & 152 & 42.8\% & 13.2\% & 7.6 \\
\bottomrule
\end{tabular*}
\begin{tablenotes}
\bmctablenotefont
\item TP = observed deaths among selected high-risk children. NNS = screened children per observed death identified. The full 5\%, 10\%, 15\%, and 20\% capacity table is provided as Additional file 4.
\end{tablenotes}
\end{table}
\end{bmcnolinenumbers}

\FloatBarrier

\subsection{Geographic Fairness Analysis: Equity Gradient}

Structural, preventable causes such as poverty and sanitation explain most of the variation in mortality in poorer regions and are measured directly in the survey, giving the model strong predictive signal. In wealthier regions such as Dhaka and Khulna, which are also more medicalised \cite{Pulok2018,Adams2019}, the basic needs for survival are largely met and residual mortality is driven by stochastic biological events (congenital anomalies, intrapartum complications, birth asphyxia) or unmeasured clinical complications that household survey instruments cannot capture.

\begin{figure}[H]
\centering
\includegraphics[width=0.95\textwidth]{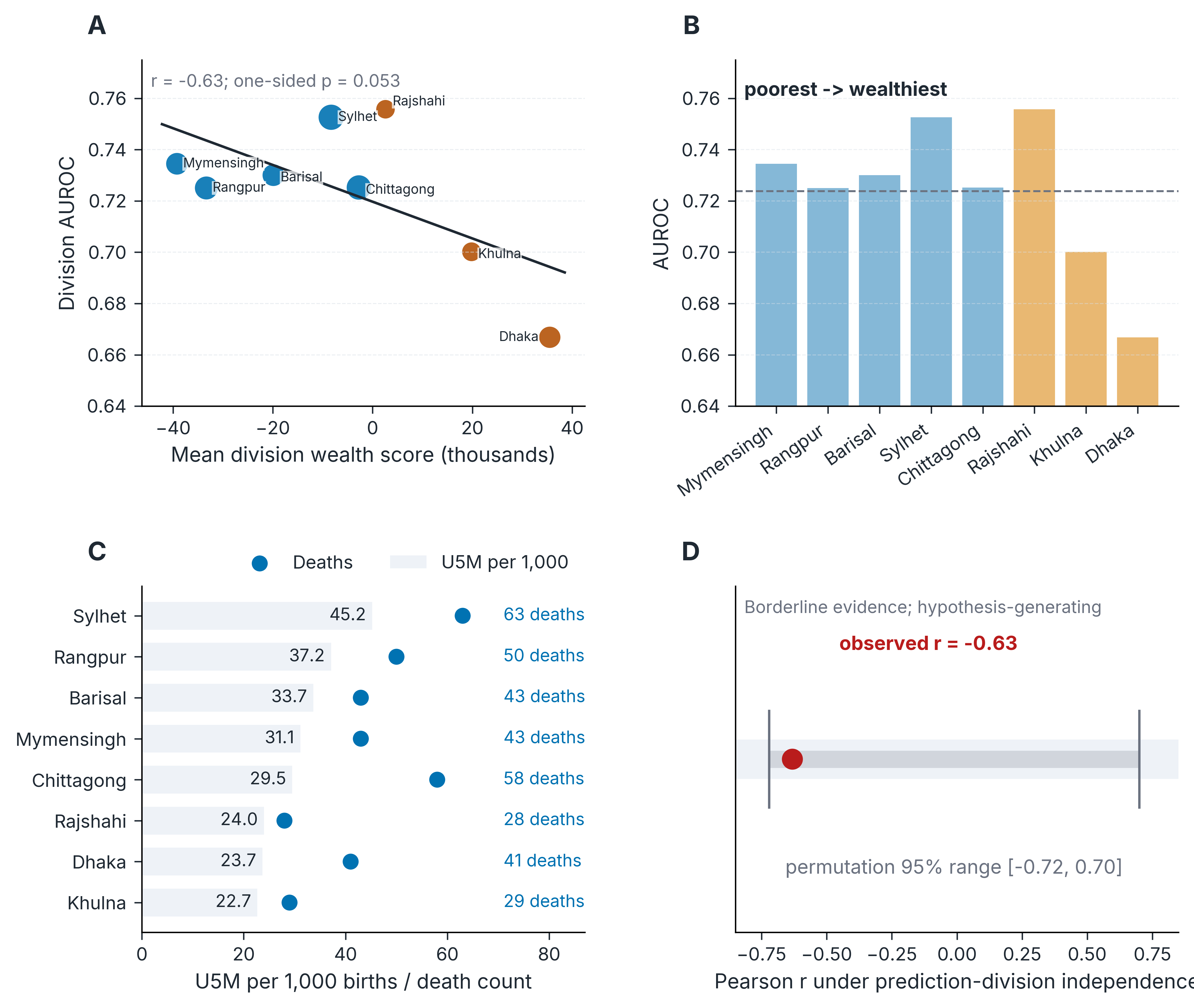}
\caption{\textbf{Geographic equity gradient under temporal validation.} Division-level wealth, AUROC, event-count context, and permutation evidence are shown together to separate the observed negative gradient from the small-stratum uncertainty behind it.}
\label{fig:equity}
\end{figure}

This pattern indicates heterogeneity in survey-based predictability across divisions; its policy interpretation is considered in the Discussion.

\begin{bmcnolinenumbers}
\begin{table}[!htbp]
\centering
\bmctablefont
\setlength{\tabcolsep}{2pt}
\caption{\textbf{Division-level context for the equity-gradient analysis.} Divisions are ordered from lowest to highest wealth rank; the table adds event-count and mortality context to the visual gradient in Figure~\ref{fig:equity}.}
\label{tab:regional}
\begin{tabular*}{\textwidth}{@{\extracolsep{\fill}}lrrrrr@{}}
\toprule
\textbf{Division} & \textbf{$n$} & \textbf{Deaths} & \textbf{U5M/1 000} & \textbf{AUC rank} & \textbf{AUROC} \\
\midrule
Mymensingh & 1381 & 43 & 31.1 & 3 & 0.735 \\
Rangpur & 1345 & 50 & 37.2 & 6 & 0.725 \\
Barisal & 1276 & 43 & 33.7 & 4 & 0.730 \\
Sylhet & 1393 & 63 & 45.2 & 2 & 0.753 \\
Chittagong & 1965 & 58 & 29.5 & 5 & 0.725 \\
Rajshahi & 1167 & 28 & 24.0 & 1 & 0.756 \\
Khulna & 1280 & 29 & 22.7 & 7 & 0.700 \\
Dhaka & 1731 & 41 & 23.7 & 8 & 0.667 \\
\bottomrule
\end{tabular*}
\begin{tablenotes}
\bmctablenotefont
\item Rows are ordered by wealth rank, from the poorest division to the wealthiest. Permutation summary: Pearson $r = -0.632$, one-sided permutation $p = 0.053$, two-sided $p = 0.099$, null 95\% range -0.72 to 0.70.
\item AUROC values are shown only as contextual detail; Figure~\ref{fig:equity} remains the primary visual display of the wealth-performance gradient.
\end{tablenotes}
\end{table}
\end{bmcnolinenumbers}

A division with a higher wealth score has greater wealth. Division-level AUROC exceeded 0.72 in five of the six poorer or higher-mortality divisions and was lower in Dhaka (0.667) and Khulna (0.700). The Pearson correlation between division mean wealth and division AUROC was $r = -0.63$ ($p = 0.09$, $n = 8$ divisions) under the 2022 test-period wealth convention, and $r = -0.75$ ($p = 0.03$) under an all-study-period wealth convention as a sensitivity.

Because the test has $n = 8$ divisions and unequal event counts (28--63 deaths per division), we additionally ran a permutation null: holding division labels, division event counts, and division mean wealth fixed, we randomly permuted the saved NAS predicted probabilities across the 11 538 2022 test rows 5 000 times, recomputing the division-level AUROCs and the wealth-AUROC Pearson $r$ each time. The null distribution of $r$ was symmetric about zero with a 95\% range of $[-0.72, +0.70]$, reflecting genuine low power at $n = 8$. The observed $r = -0.63$ fell in the negative tail of this null (one-sided $p = 0.053$, two-sided $p = 0.099$), so the gradient is borderline-significant rather than convincingly demonstrated. We therefore treat the equity gradient as a hypothesis-generating observation whose direction is consistent across both wealth conventions and survives an event-count-equalised permutation, but whose definitive demonstration requires either replication across DHS countries or finer-grain (sub-district) geographic strata.

\subsection{Model Calibration and Sensitivity Analyses}

\begin{figure}[H]
\centering
\includegraphics[width=0.90\textwidth]{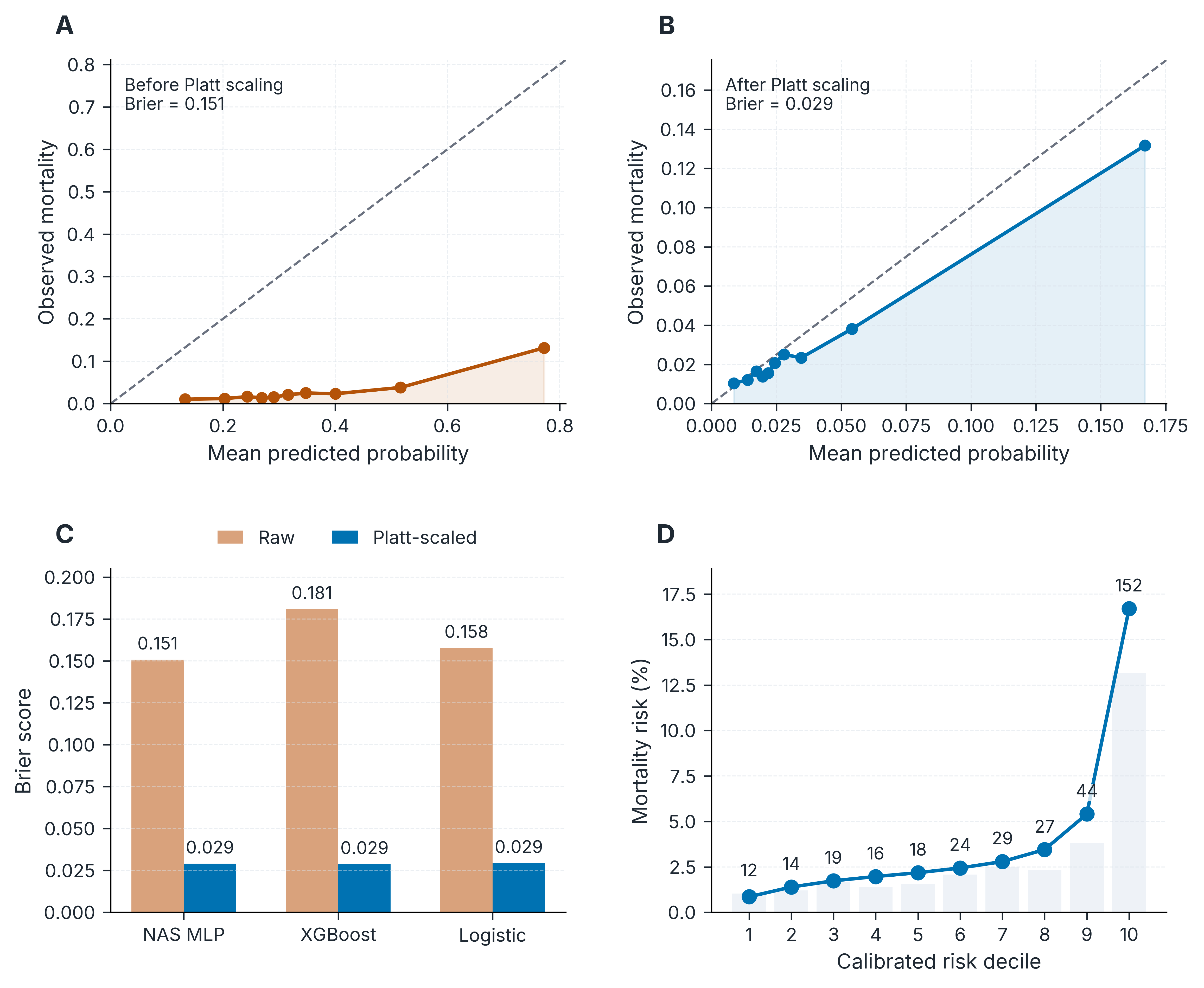}
\caption{\textbf{Reliability calibration of the temporally validated models.} Raw class-weighted probabilities are over-confident, while Platt scaling on 2017 scores produces calibrated 2022 probabilities. Brier scores and calibrated NAS risk deciles summarize the probability-facing evidence.}
\label{fig:calibration}
\end{figure}

All three models' predicted probabilities required Platt-scale recalibration before policy interpretation, because each was trained with class-weighted loss to address the 3.8\% positive-class prevalence. Platt scaling \cite{Platt1999} on 2017 reduced 2022 test-set Brier scores from 0.151 to 0.029 (NAS), 0.181 to 0.029 (XGBoost), and 0.158 to 0.029 (logistic). The Figure~\ref{fig:calibration} reliability diagrams illustrate the NAS pre/post calibration. After Platt scaling, division-level calibration was stable, with post-calibration Brier scores ranging from 0.021 (Khulna) to 0.042 (Sylhet), supporting the use of calibrated probabilities for screening-yield summaries.

To examine whether the temporal estimate was sensitive to the complex BDHS sampling design, we calculated weighted AUROC using DHS sampling weights via a design-aware bootstrap. The weighted AUROC (0.721 [95\% CI 0.693--0.752]) was statistically indistinguishable from the unweighted estimate, supporting the robustness of individual-level discrimination to survey weighting.

We also conducted a censoring-sensitivity analysis. The DHS captures children born within five years prior to the survey, so most test children were under 60 months at interview. In the Bangladesh sample, 71.0\% of observed under-five deaths occurred in the neonatal period (first 28 days) and 90.9\% in the first year; Li et al.\ 2021 \cite{Li2021} report 53.1\% of LMIC under-five deaths in the neonatal period as a global benchmark. Stratifying the GA-v2 NAS predictions by minimum age at interview (alive children only), test-set AUROC was 0.730 [0.700, 0.758] at no restriction, 0.719 [0.688, 0.748] at $\geq$12 months observed, and 0.706 [0.676, 0.735] at $\geq$24 months observed---restrictions up to and including 36 months remain within bootstrap confidence-interval overlap of the primary estimate. At the 48-month restriction the AUROC drops to 0.662 [0.628, 0.695] (n = 1 256, with only 901 alive children retained), which does \emph{not} overlap with the primary; we therefore treat censoring as a substantive limitation in settings where prediction is intended for survival beyond the first year of life (see Limitations). The age-at-death distribution, a cumulative incidence curve, and the full stratified-AUROC table are reported in Additional file 2: Table S4 and Additional file 3: Figure S2.

Because the policy-facing summaries depend on the trade-off between true positives and false positives at a given threshold, we also computed Vickers-Elkin Decision Curve Analysis on the Platt-calibrated temporal predictions (Additional file 3: Figure S4; Table~\ref{tab:calibration_utility}) \cite{Vickers2006Decision}. All three model classes yielded positive net benefit relative to ``treat-none'' and ``treat-all'' across threshold probabilities of policy interest (1\%--15\%), with curves overlapping within plotting resolution. At the 5\% threshold (the calibrated probability corresponding to roughly the top-10\% screening capacity), net benefit was approximately 0.0083 for all three models, equivalent to about 8 additional true deaths detected per 1 000 children screened over a treat-none baseline; at this threshold, ``treat-all'' was negative (about $-0.020$) and ``treat-none'' was zero. The DCA supports the regression and screening-yield findings: model architecture is interchangeable within the standardised pipeline, and the validation regime determines what the displayed net benefit represents.

\begin{bmcnolinenumbers}
\begin{table}[!htbp]
\centering
\bmctablefont
\setlength{\tabcolsep}{3pt}
\caption{\textbf{Calibrated probability and decision-utility summary.} Brier scores and net benefit are reported for the Platt-calibrated temporal predictions.}
\label{tab:calibration_utility}
\begin{tabular*}{\textwidth}{@{\extracolsep{\fill}}lrrrrrr@{}}
\toprule
\textbf{Model} & \textbf{$n_{\mathrm{test}}$} & \textbf{Deaths} & \textbf{Raw Brier} & \textbf{Platt Brier} & \textbf{NB/1 000 1\%} & \textbf{NB/1 000 5\%} \\
\midrule
NAS MLP & 11538 & 355 & 0.151 & 0.029 & 21.3 & 8.3 \\
XGBoost & 11538 & 355 & 0.181 & 0.029 & 20.9 & 8.3 \\
Logistic & 11538 & 355 & 0.158 & 0.029 & 21.0 & 8.2 \\
\bottomrule
\end{tabular*}
\begin{tablenotes}
\bmctablenotefont
\item NB/1 000 is net benefit multiplied by 1 000 children at the specified calibrated risk threshold. Treat-none has net benefit 0 by definition; treat-all is inferior at the 5\% threshold.
\item Additional file 2: Table S3 provides the 10\% and 15\% threshold values.
\end{tablenotes}
\end{table}
\end{bmcnolinenumbers}

\begin{figure}[H]
\centering
\includegraphics[width=0.95\textwidth]{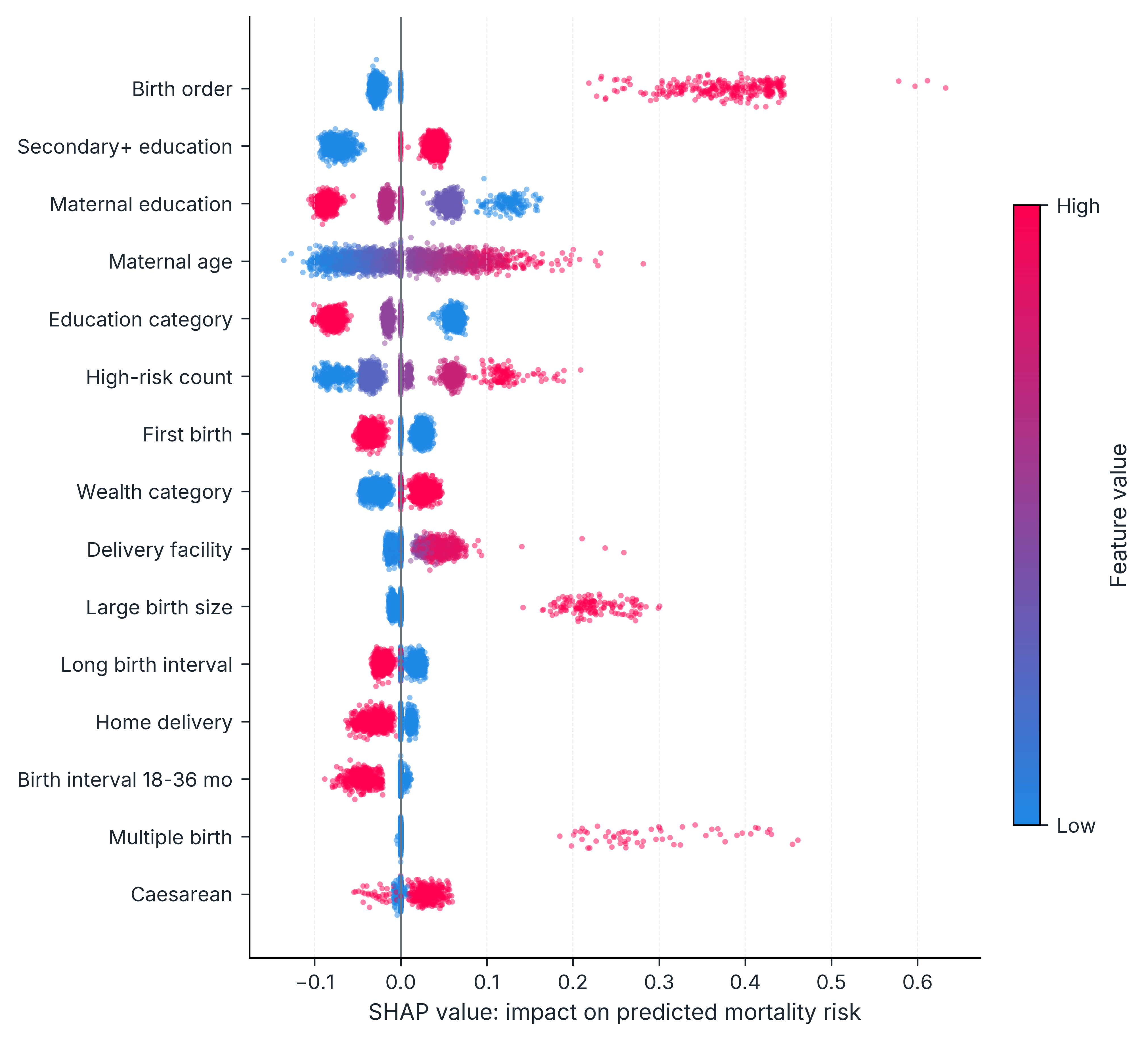}
\caption{\textbf{SHAP feature-impact distribution for temporally validated NAS predictions.} Each row shows the distribution of SHAP values across 3 000 randomly sampled 2022 test-set children for the top-ranked predictors; points are colored from blue (lower within-feature values) to pink/red (higher within-feature values) so direction and heterogeneity can be read without adding a separate tabular SHAP panel.}
\label{fig:shap_explainability}
\end{figure}

\FloatBarrier
SHAP analysis \cite{Lundberg2017} demonstrated that the NAS model had learned biologically plausible relationships consistent with decades of child-survival research (Figure~\ref{fig:shap_explainability}).

The leading predictors reflect an interaction between biological susceptibility and socioeconomic context. Birth order (mean $|$SHAP$|$ = 0.062) identifies a U-shaped risk curve: firstborns face physiological problems of primiparity, while higher-order births face maternal depletion and resource dilution. Education-related features follow next: secondary-plus education (0.050), broader maternal education (0.045), and categorical education level (0.040), consistent with a protective socioeconomic gradient in the beeswarm. Maternal age (0.044) captures elevated risk at vulnerable age extremes, especially adolescent motherhood. The next tier includes the high-risk-count summary (0.035), which captures co-occurring vulnerabilities such as young age, short interval, and high parity.

\subsection{Model Ablation and Prospective Generalization}

To characterise the predictive drivers and architectural requirements of the model under prospective domain shift, we conducted an ablation study evaluating (i) public-health feature domains and (ii) neural-network hyperparameter choices, all under the strict five-year prospective temporal regime (train 2011--2014, validation 2017, test 2022).

\subsubsection{Feature-category ablation}

\begin{figure}[!htbp]
\centering
\includegraphics[width=0.95\textwidth]{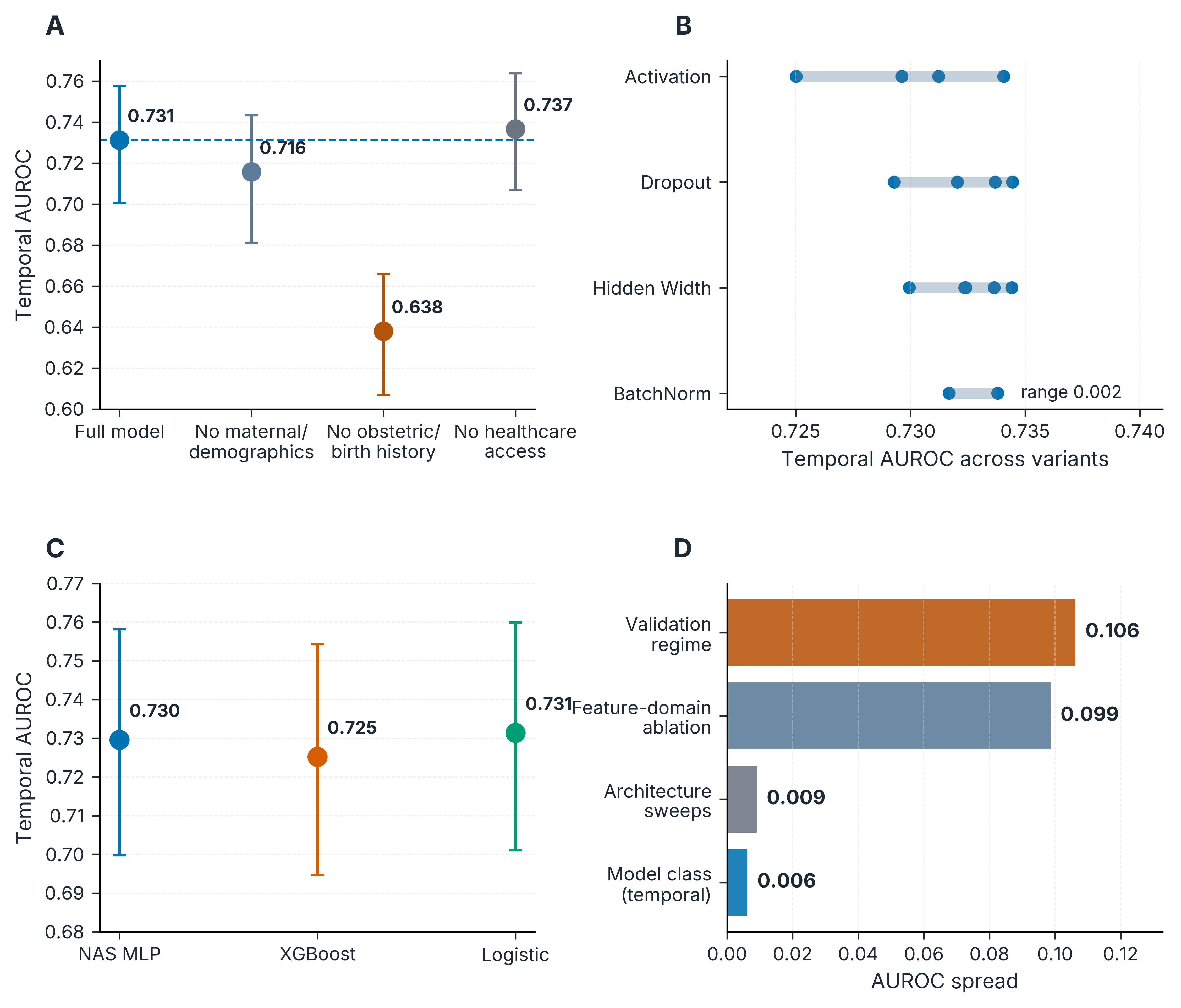}
\caption{\textbf{Ablation and generalisation scale under prospective temporal validation.} Feature-domain ablation, hyperparameter sweeps, model-class comparison, and effect-size summaries show that validation-regime choice is larger than architecture-level variation.}
\label{fig:feature_group_ablation}
\end{figure}

We grouped the 26 model features into three public-health domains: demographics (10), obstetric and birth history (12), and healthcare access (4). The full model achieved temporal test AUROC 0.730 [0.700, 0.758] (Figure~\ref{fig:feature_group_ablation}).

\begin{itemize}
\item \textbf{Obstetric and birth history.} Removing this domain dropped prospective discrimination to 0.638 [0.607, 0.666] (deficit $-$0.093), identifying previous birth intervals, birth order, parity, and obstetric-risk markers as the core biological drivers of predictive capacity.
\item \textbf{Maternal and household demographics.} Removing demographic variables (maternal age, education, residence, wealth) moderately reduced AUROC to 0.716 [0.681, 0.743] (deficit $-$0.015), confirming their role as critical contextual covariates.
\item \textbf{Healthcare access.} Surprisingly, ablating all healthcare-access features (facility delivery, caesarean, ANC visits, skilled birth attendance) yielded a stable or slightly improved temporal AUROC of 0.737 [0.707, 0.764]. Under strict prospective temporal validation, reported healthcare-access indicators do not add independent predictive power, likely due to socioeconomic confounding and structural shifts in service delivery between survey rounds.
\end{itemize}

\subsubsection{Hyperparameter sweeps}

Prospective AUROC was robust to activation choice (ELU 0.730, ReLU 0.725, sigmoid 0.734, identity/linear 0.731), with a maximum spread of 0.009. Regularisation provided modest but consistent generalisation gains: dropout rates of 0.1 / 0.3 / 0.5 gave AUROCs of 0.732 / 0.734 / 0.735, vs.\ 0.729 without dropout. Hidden-layer width followed an inverted-U pattern across widths 8/16/32/64/128, with AUROCs 0.732/0.734/0.734/0.732/0.730, confirming that narrower networks generalise at least as well as wider ones (Additional file 3: Figure S1). Omitting batch normalisation (0.734) slightly outperformed including it (0.732), consistent with the GA's discovery of the no-BatchNorm architecture.

These ablation results reinforce our core thesis. Validation-regime choice dominates minor architectural details. Across 13 architectural variants, prospective AUROC spans only 0.006--0.009. This is an order of magnitude smaller than the validation-regime effect (0.09--0.11).

\section{Discussion}

The principal finding is that validation-regime choice changed the public-health interpretation of child-mortality prediction more than model architecture did. Under the standardised benchmark, the same GA-v2 MLP varied from AUROC 0.669 under 2022-only random validation to 0.775 under pooled random validation, while the cross-survey temporal estimate was 0.730. Within any validation regime, the spread between the GA-v2 MLP, XGBoost, and logistic regression was at most 0.015. This study should therefore not be read as evidence that neural architecture search is superior to gradient-boosted trees or logistic regression for tabular DHS data. It shows instead that validation design determines whether a reported screening result can support prospective programme planning.

Both directions of validation bias matter for policy. Pooled random validation over-estimated temporal AUROC and made screening appear more efficient than it would be prospectively. Single-round 2022 random validation moved in the opposite direction, under-estimating temporal performance because it trained on a smaller and narrower sample. This is relevant to DHS-machine-learning studies because both practices are common: pooled analyses can create overconfident expectations about deployment, while single-round random analyses can make a potentially useful screening approach look less efficient than it is under cross-round temporal validation.

The equity gradient requires interpretation against the algorithmic-fairness literature. Vyas, Eisenstein, and Jones \cite{Vyas2020} critique clinical algorithms that take race as an explicit input variable, arguing that race correction entrenches structural inequity in downstream care. Obermeyer and colleagues \cite{Obermeyer2019} demonstrate how a model using healthcare spending as a proxy for healthcare need systematically under-prioritises Black patients, because unequal historical spending encodes structural racism in the training label. Two features of our setting differ: first, our model takes neither race nor ethnicity as input, and division is a geographic-economic indicator in an ethnically homogeneous country, not a racial category; second, our outcome (death before age five recorded in the BDHS) is a terminal event observed identically across socioeconomic groups, not a proxy that itself encodes upstream inequity. The mechanism behind our division-level AUROC gradient is therefore distinct: it reflects how much variance in under-five mortality is captured by household-survey predictors in each setting, which itself reflects the stage of the local epidemiologic transition \cite{Omran2005}.

These observations qualify, rather than dismiss, the fairness frameworks of Vyas and Obermeyer. Where performance disparity arises from biased inputs or biased proxy labels, equal performance across groups is an appropriate equity criterion. Where performance disparity arises from genuine differences in how much variance in the outcome is structurally captured by available predictors---as in our cross-divisional comparison---uniform AUROC is neither expected nor a meaningful goal. Dhaka's lower AUROC reflects the fact that under-five mortality there is increasingly driven by stochastic biological events that are not visible to a household-survey instrument, not algorithmic failure.

\subsection{Public Health and Policy Implications}

Under-five mortality is used as an SDG child-survival indicator and reflects social, economic, environmental, and health-system conditions \cite{Ahmed2025}. Random-split AUROC should not be used alone to guide procurement, screening programmes, or resource allocation, because it can misstate the future screening yield that a programme should expect. Temporal NNS and sensitivity at a fixed screening capacity are more actionable: they tell programme planners how many children must be followed up to identify one observed death and what share of deaths would be captured if the highest-risk 10\% were prioritised.

Under cross-survey temporal validation, the GA-v2 MLP identified 152 of 355 deaths in the 2022 test set at the top-10\% threshold (sensitivity 42.8\%, PPV 13.2\%, NNS 7.6). Pooled random validation would have suggested a more optimistic NNS of 5.6; 2022-only random validation would have suggested a more pessimistic NNS of 11.0. Either estimate could mislead programme planning. The temporal estimate is the defensible basis for screening workload, community-health-worker follow-up, referral planning, and budget scenarios. In practical terms, a top-10\% screening programme based on the temporally validated model would need follow-up for about 1 154 children in the 2022 test population to identify 152 observed deaths.

The geographic gradient also affects how such a model should be used. In poorer, higher-mortality divisions, household-survey predictors carry stronger signal, so risk stratification may help prioritise community-health-worker follow-up, nutrition support, postnatal contact, and referral. In lower-mortality urban settings such as Dhaka, lower predictability should not be treated simply as algorithmic failure; it suggests that a larger share of residual mortality may be driven by causes less visible to household surveys, including congenital anomaly, intrapartum complications, birth asphyxia, and facility-level quality. For those settings, policy response should emphasise health-system strengthening and quality of care, not only individual risk scoring.

\subsection{Limitations}

Several limitations deserve acknowledgment. The DHS captures children born within 5 years prior to the survey, so most test children were less than 60 months old at interview, creating incomplete theoretical follow-up for under-five mortality. Discrimination is robust to censoring restrictions of up to 36 months at interview (AUROC remains within bootstrap CI overlap of the primary), but the restriction to children observed for the full 48 months yields a smaller sample (n = 1 256) and a non-overlapping AUROC of 0.66 (Additional file 2: Table S4). Censoring therefore remains a substantive limitation for any deployment that targets late-childhood mortality after infancy.

Missing-data rates were similar across divisions for several healthcare variables, but missingness was associated with mortality for some variables. The tested imputation strategies did not materially change temporal XGBoost discrimination, but multiple imputation with chained equations should be considered in future work. Features such as birth size and delivery mode are captured post-birth; therefore, deployment is restricted to postpartum screening rather than prenatal prediction. External validity beyond Bangladesh is not established: the findings reflect Bangladesh's specific demographic and health-system context, DHS variable continuity across the 2011/2014/2017/2022 rounds, and marked division-level heterogeneity. Generalisation to other LMIC contexts requires explicit cross-country temporal validation. Finally, the observational nature of DHS data prevents causal interpretation of predictor effects.

\section{Conclusions}

In a four-round Bangladesh DHS benchmark, validation regime changed the apparent screening efficiency and policy value of under-five mortality prediction more than model architecture did. Pooled random validation made screening appear more efficient than it was under temporal testing; 2022-only random validation made it appear less efficient. Cross-survey-round temporal validation therefore provides the most policy-relevant estimate for future deployment. At a fixed top-10\% screening capacity, the temporal GA-v2 MLP identified 42.8\% of observed deaths with NNS 7.6, statistically indistinguishable from XGBoost and logistic regression. The public-health contribution of this work is not neural-network superiority; it is a reproducible demonstration that temporal validation and deployment-facing metrics are necessary before DHS-based prediction models are used for child-health planning. Multi-country temporal replication and linkage to intervention pathways are the next steps.

\vspace{0.5em}

\backmatter

\bmhead{List of abbreviations}
ANC, antenatal care; AUROC, area under the receiver operating characteristic curve; BDHS, Bangladesh Demographic and Health Survey; CI, confidence interval; DCA, decision curve analysis; DHS, Demographic and Health Survey; EA, enumeration area; GA, genetic algorithm; MLP, multilayer perceptron; NAS, neural architecture search; NB, net benefit; NNS, number needed to screen; PPV, positive predictive value; ROC, receiver operating characteristic; SHAP, Shapley additive explanations; TRIPOD+AI, Transparent Reporting of a multivariable prediction model for Individual Prognosis Or Diagnosis plus Artificial Intelligence; U5M, under-five mortality.

\bmhead{Supplementary information}
Supplementary tables, supplementary figures, the completed TRIPOD+AI checklist, source data for the main tables, and a combined supplementary PDF accompany this manuscript as additional files.

\bmhead{Additional files}
\noindent\textbf{Additional file 1.} PDF. Completed TRIPOD+AI checklist for the prediction-model reporting items.\\
\textbf{Additional file 2.} TEX and CSV. Supplementary tables S1--S7 supporting the validation-regime, paired-bootstrap, calibration, censoring, ablation, missing-data, and SHAP analyses.\\
\textbf{Additional file 3.} PNG and PDF. Supplementary figures S1--S6.\\
\textbf{Additional file 4.} CSV. Source data for the main manuscript tables.\\
\textbf{Additional file 5.} PDF. Combined supplementary material with manuscript title, author names, table of contents, Supplementary Tables S1--S7, and Supplementary Figures S1--S6.

\section*{Declarations}

\bmhead{Ethics approval and consent to participate}
This study used de-identified public-use Bangladesh Demographic and Health Survey datasets obtained through the DHS Program after registration and authorization. The original BDHS survey protocols were reviewed by the relevant national ethics authority in Bangladesh and the ICF Institutional Review Board, and informed consent was obtained by the survey implementers. The present secondary analysis involved no new participant recruitment, contact, intervention, or identifiable human data.

\bmhead{Consent for publication}
Not applicable.

\bmhead{Availability of data and materials}
The BDHS datasets analysed during the current study are available from the DHS Program after registration and approval \cite{DHSProgram2026}. Table source data and supplementary materials are included in the additional files accompanying this manuscript.

\bmhead{Competing interests}
The authors declare that they have no competing interests.

\bmhead{Funding}
No external funding was received for this study.

\bmhead{Authors' contributions}
MMMF conceptualised the study, prepared datasets, implemented modelling and validation, produced tables and figures, and drafted the manuscript. MMH and MS contributed to methodology, interpretation, and critical manuscript revision. MRK supervised the study, contributed to methodological and statistical interpretation, and revised the manuscript critically. All authors read and approved the final manuscript.

\bmhead{Acknowledgements}
The authors acknowledge the DHS Program, the National Institute of Population Research and Training, and ICF for access to the Bangladesh Demographic and Health Survey datasets.

\bmhead{Authors' information}
Not applicable.

\bibliography{references}

\end{document}